\documentclass{article}

\usepackage{microtype}
\usepackage{graphicx}
\usepackage{subfigure}
\usepackage{booktabs} 

\usepackage{hyperref}

\usepackage[accepted]{icml2024}

\usepackage{amsmath}
\usepackage{amssymb}
\usepackage{mathtools}
\usepackage{amsthm}

\usepackage[capitalize,noabbrev]{cleveref}

\theoremstyle{plain}

\theoremstyle{definition}

\theoremstyle{remark}

\usepackage[textsize=tiny]{todonotes}

\icmltitlerunning{Accepted at the ICML 2024 Workshop on AI for Science}

\begin{document}

\twocolumn[
\icmltitle{An Advanced Physics-Informed Neural Operator for Comprehensive Design Optimization of Highly-Nonlinear Systems: An Aerospace Composites Processing Case Study}




\begin{icmlauthorlist}
\icmlauthor{Milad Ramezankhani}{yyy}
\icmlauthor{Anirudh Deodhar}{yyy} 
\icmlauthor{Rishi Yash Parekh}{yyy}
\icmlauthor{Dagnachew Birru}{yyy}



\end{icmlauthorlist}

\icmlaffiliation{yyy}{Quantiphi, Marlborough, MA 01752, United States of America}

\icmlcorrespondingauthor{Milad Ramezankhani}{milad.ramezankhani@quantiphi.com}

\icmlkeywords{Machine Learning, ICML}

\vskip 0.3in
]



\printAffiliationsAndNotice{}  

\begin{abstract}
Deep Operator Networks (DeepONet) and their physics-informed variants have shown significant promise in learning mappings between function spaces of partial differential equations (PDEs), enhancing the generalization of traditional neural networks. However, for highly nonlinear real-world applications like aerospace composites processing, existing models often fail to capture underlying solutions accurately and are typically limited to single input functions, constraining rapid process design development. This paper introduces an advanced physics-informed DeepONet tailored for such highly nonlinear systems with multiple input functions. Equipped with architectural enhancements like nonlinear decoders and effective training strategies such as curriculum learning and domain decomposition, the proposed model handles high-dimensional design spaces with significantly improved accuracy, outperforming the vanilla physics-informed DeepONet by two orders of magnitude. Its zero-shot prediction capability across a broad design space makes it a powerful tool for accelerating composites process design and optimization, with potential applications in other engineering fields characterized by strong nonlinearity.
\end{abstract}

\section{Introduction}
\label{submission}

The simulation and optimization of engineering and scientific systems involves solving a set of partial differential equations (PDEs) across a range of system parameters. The study of these parametric PDEs involves obtaining the solution under different input functions such as initial condition (IC), boundary conditions (BCs), source functions, geometries, and coefficients. This entails repeated execution of computationally expensive numerical solvers such as finite element/volume methods (FEM/FVM) \cite{zauderer2011partial}. Reduced-order methods have been introduced to enhance computational efficiency at the expense of reduced accuracy, offering a more practical solution for exploring the design space of parametric PDEs \cite{lucia2004reduced}. Similarly, recent advancements in deep learning, specifically in the field of scientific machine learning (SciML) empower the use of ML models for facilitating scientific discovery and optimization \cite{cuomo2022scientific}. SciML models infuse the unparallel approximation capabilities of ML models with the established governing physical laws to primarily accomplish one of the following tasks: 1) solve PDEs, 2) discover PDE parameters, and 3) learn operators. Methods such as physics-informed neural networks (PINNs) \cite{raissi2019physics} and deep Galerkin method \cite{sirignano2018dgm} have been introduced to approximate the solution of PDEs, however, they are inherently problem-specific and thus require retraining/finetuning in order to adapt to new system configurations. Although some enhancements have been introduced to improve their generalizability \cite{jagtap2020extended,gao2021phygeonet, ramezankhani2023sequential, majumdar2024hxpinn}, they still lack the adaptability required for the optimization tasks. Operator learning on the other hand targets the discovery of an unknown mathematical operator governing a PDE system \cite{boulle2023mathematical}. It seeks to capture a nonlinear mapping from one space of functions (inputs) to another space of functions (outputs). For a physical system described by PDEs, the input functions typically are the initial condition \(u_0(x)\), boundary conditions \(u_{bc}(t, x)\), and forcing term \(f(t, x)\). The solution of PDE \(u(t, x)\) is considered as the output function \cite{lu2022comprehensive}.

The operator is typically represented by a \textit{neural operator}, a generalized form of neural networks, which can take functions (in a discretized form) as the inputs and outputs \cite{boulle2023mathematical}. Unlike PINNs, neural operators offer real-time prediction capabilities for varying system configurations without the need for retraining. This makes the neural operator an ideal tool for performing parametric PDE studies. The training of neural operators, however, is computationally expensive and can incur poor generalization performance. Hence, several architectures have been introduced to tackle such shortcomings, including deep neural operator (DeepONet) \cite{lu2021learning}, Fourier neural operator (FNO) \cite{li2020fourier}, Graph neural operator \cite{li2020neural} and Wavelet Neural Operator (WNO) \cite{tripura2022wavelet},  and deep Green network \cite{gin2021deepgreen}. These models are different in terms of discretization approach as well as approximation techniques employed for efficiency and scalability.

DeepONet, a neural operator method theoretically motivated by the universal operator approximation theorem \cite{chen1995universal}, offers a more generalized operator learning framework to discover nonlinear function mappings. The two-part architecture of DeepONet consists of a branch net responsible for distilling the operator’s input function into a fixed-size latent vector, and a trunk net, which decodes the output of the branch net to generate the final output at the specified locations. While DeepONet is primarily developed as a fully data-driven method, physical governing laws can also be incorporated to learn the solution operators in a fully physics-informed (i.e., data-agnostic) manner. Specifically, the existing governing equations and physical laws are integrated into the training of the DeepONet as additional loss terms \cite{wang2021learning}. Despite the success of DeepONet and its physics-informed variant in effectively learning operators for a range of benchmark problems, it has been shown that the existing architectures may struggle to accurately capture the dynamics of PDE systems in complex real-life scenarios. In particular, for nonlinear submanifolds in function spaces, the finite-dimensional linear representation of DeepONet’s decoder might be insufficient to learn the true target functions \cite{seidman2022nomad}. They also fall short when trained against multiphysics problems with coupled PDEs due to interactions between system variables, intricate geometries and lack of sufficient training data \cite{rahman2024pretraining}. Furthermore, the physics-informed DeepONet (PIDON) also shares the same shortfalls as PINNs \cite{wang2022and}. Among them are failure to learn the long temporal domain, reduced accuracy against sharp edges and nonlinearities, and poor performance in multi-scale and multi-physics problems \cite{krishnapriyan2021characterizing}. 

This paper introduces an enhanced PIDON architecture capable of learning the solution operator mapping multiple input functions (i.e., design variables) to multiple output functions (i.e., decision variables) in a complex, highly nonlinear, and multi-physics engineering problem with a long temporal domain. Specifically, we explore the data-agnostic learning of the solution operator associated with coupled PDEs governing the thermochemical behavior of the aerospace-grade composites curing process in an autoclave under various input functions (i.e., process and material design configurations). Our investigation reveals the failure of DeepONets’ original architecture to capture the complex dynamics inherent in the problem domain. To address this, we introduce a series of architectural improvements by integrating nonlinear decoders and employing multiple branch networks as well as leveraging advanced learning techniques such as curriculum learning and domain decomposition. We show that the proposed enhanced PIDON architecture can successfully learn the solution operator for this highly nonlinear PDE system and generate accurate predictions across various system configurations. While previous attempts have explored the application of data-driven neural operators \cite{chen2021residual, rashid2022learning, chen2023physics}  and recently physics-informed FNO \cite{meng2023novel} in composites processing, their scope has been largely limited to singular design variables. Our work, on the other hand, broadens the horizon by incorporating multiple design parameters including but not limited to the cure cycle recipe, heat transfer coefficients (HTCs), and material thickness. This enhances the generalizability of neural operators, paving the path toward building scientific foundation models for more comprehensive and full-scale modeling and design optimization of complex engineering scenarios. Furthermore, we show that our enhanced PIDON outperforms previous models in terms of predictive performance. 

The proposed framework refines the original DeepONet architecture to better tackle the inherent complexities present in real-world engineering scenarios, thereby enhancing its applicability and effectiveness. The contributions of this paper can be summarized as follows:
\begin{itemize}
  \setlength{\itemsep}{2pt}
  \setlength{\parskip}{0pt}
  \setlength{\parsep}{0pt}
    \item Introduce an enhanced PIDON framework, featuring nonlinear decoders and multiple branch networks, to account for high nonlinearity and diverse input functions in complex PDE systems;
    \item Investigate the effectiveness of a series of remedies, such as curriculum learning and domain decomposition, typically used for efficient learning of PINNs, on the performance of PIDON;
    \item Develop a customized neural operator framework via introducing local spatial coordinates, enabling seamless learning of solution operators for the thermochemical curing process of composites in autoclave across multiple process design variables.
\end{itemize}

\subsection{Application to advanced composites manufacturing} \label{compositessection}

Superior mechanical properties, light weight and high durability have made fiber-reinforced polymer composites a popular choice of materials in high-performance applications where very large unified structures with intricate geometries are manufactured. Specifically in autoclave processing (Figure \ref{Curecycle_schematic}), a system of resin-impregnated fibers and a tool is placed in an autoclave and subjected to a predefined temperature and pressure cycle (i.e., cure cycle) \cite{strong2008fundamentals}. The objective of this process is to cure the resin system in a way to achieve a uniform resin cure, optimum resin content, and a void-free product while minimizing any process-induced residual stress and deformation \cite{hubert2001cure}. The through-thickness temperature profiles in the part and tooling as well as the evolution of the resin degree of cure are the key state variables of the composite system and crucial for foreseeing the process-induced defects.  For any new scenario including modifications in part geometry and material properties, a process design optimization needs to be carried out via modifying a baseline cure cycle as well as optimizing the tooling and structure designs to ensure the above criteria are met. This iterative trial-and-error procedure is very expensive and time-consuming, especially for large structures, and thus prohibitive for real-world applications. Computational models have been developed as a more efficient alternative to model various aspects of the curing process such as heat transfer, resin cure kinetics, and residual stress development \cite{van2009hexply}. While numerical methods such as FEM provide accurate approximations to the solution of PDEs, they can easily become computationally prohibitive when it comes to the iterative procedure of curing process optimization. The proposed PIDON model, on the other hand, can generate real-time predictions for various process configurations, significantly expediting the design optimization procedure. This enables a much faster exploration of the design space (e.g., processing scenarios, part and tooling designs) to identify the optimum process configurations for obtaining the desired material properties. In particular, following the terminology developed in \cite{fabris2018framework}, for a composites manufacturing system with four main constitutive components (i.e., Process, Tools, Equipment and Parts), the following design variables can be handled by the proposed PIDON model (Figure \ref{Curecycle_schematic}):
\begin{itemize}
  \setlength{\itemsep}{2pt}
  \setlength{\parskip}{0pt}
  \setlength{\parsep}{0pt}
    \item Process: cure cycle specifications including heating rate ($r_1$ and $r_2$), hold duration ($hd_1$ and $hd_2$), and hold temperature ($ht_1$ and $ht_2$)
    \item Tools and consumables: tool thickness ($L_t$)
    \item Equipment: convective HTCs in the autoclave ($h_{top}$ and $h_{bot}$)
    \item Parts: composite part thickness ($L_c$)
\end{itemize}

\begin{figure*}[t]
\vskip 0.2in
\begin{center}
\centerline{\includegraphics[width=0.9\textwidth]{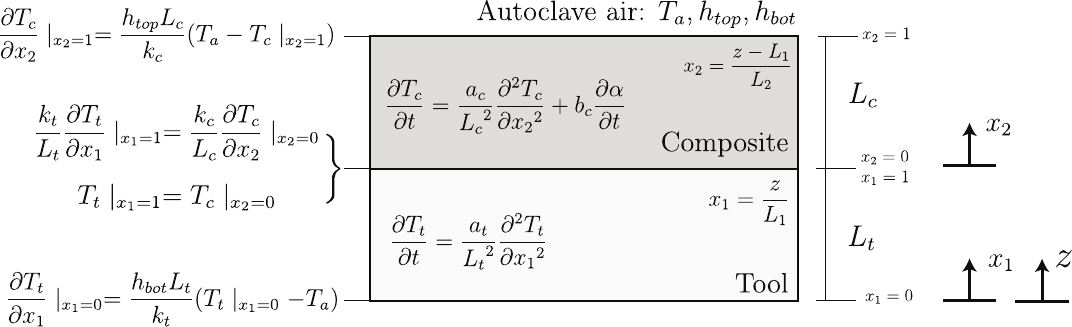}}
\caption{Schematic of composite-tool system in an autoclave with local coordinates \(x_1\) and \(x_2\).}
\label{localcoord}
\end{center}
\vskip -0.2in
\end{figure*}

The parameters concerning Parts are typically predetermined in practice according to the application requirements. During the design optimization phase, the design engineer can conveniently select and fix the values of such variables and optimize the remaining design variables accordingly. Since the part and tool thicknesses are considered among the input functions (design variables) in the training of the neural operator and vary for each prediction task, it results in inconsistencies in the system dimensions as well as the composite-tool interface location (key for satisfying the continuity condition). In order to represent all training and test cases in a unified learning framework, we introduce two local spatial coordinates that independently describe the length of each material (Figure \ref{localcoord}). The details regarding the governing equations of the thermochemical curing process as well as the implementation of local spatial coordinates are presented in \ref{CureEquations} and \ref{LocalCoordinates}.

The rest of the paper is structured as follows. Section \ref{method} presents the proposed framework and discusses the constituents of the PIDON architecture. Section \ref{results} is dedicated to an in-depth discussion of the results, investigating the performance of the proposed PIDON against the composites autoclave processing case study. Finally, the Conclusions section provides a summary and outlines future research directions.

\section{Methodology} \label{method}

\subsection{Physics-informed DeepONet}

DeepONet's architecture is composed of a branch net and a trunk net. The branch net takes the sensor point evaluations \( u = [u(x_1), u(x_2), \ldots, u(x_m)] \) as input and produces a finite-dimensional feature representation \( b = [b_1, b_2, \ldots, b_q]^T \in \mathbb{R}^q \) as output. Similarly, the trunk net encodes the inputs of the PDE system \( y \) to a feature embedding \( t = [t_1, t_2, \ldots, t_q]^T \in \mathbb{R}^q \) with the same size as the branch net's output. The output of the branch and trunk nets is then combined to calculate DeepONet's output using an element-wise product operation followed by a summation\(G_\theta(u)(y) = \sum_{k=1}^{q} b_kt_k + b_0\). In a supervised learning fashion, the DeepONet can be trained by
minimizing the error between the model’s predicted output and the actual operator solution across a range of training input functions. Appendix \ref{deeponetAppx} provides a detailed explanation of DeepONet's architecture along with a visual representation.

Drawing inspiration from PINNs, which learn the solutions of PDE systems by penalizing the residuals of the governing equations, a parallel approach is adopted in the development of PIDON framework \cite{wang2021learning}. Specifically, the output of DeepONet is constrained to align with the governing equations through the minimization of the loss function \(\mathcal{L}(\theta) = \mathcal{L}_{IC}(\theta) + \mathcal{L}_{BC}(\theta) + \mathcal{L}_{physics}(\theta)\), where \(L_{IC}\) and \(L_{BC}\) are the IC and BC losses. Assuming a constant initial condition and a Robin boundary condition, \(L_{IC}\) and \(L_{BC}\) become
\begin{equation}
\mathcal{L}_{IC}(\theta) = \frac{1}{NQ_{ic}}\sum_{i=1}^{N}\sum_{j=1}^{Q_{ic}}|G_{\theta}(u^{(i)})(y^{(i)}_j))-s^{(i)}(y^{(i)}_j)|^2
\end{equation}

\begin{align}
\mathcal{L}_{BC}(\theta) = & \frac{1}{NQ_{bc}} \sum_{i=1}^{N} \sum_{j=1}^{Q_{bc}}
\big|\alpha G_{\theta}(u^{(i)})(y^{(i)}_j) \notag \\
& + \beta\nabla G_{\theta}(u^{(i)})(y^{(i)}_j)  -\gamma \big|^2
\end{align}

where \(u^{(i)}\) is the \(i\)-th input function, \(y_j^{(i)}\) is the \(j\)-th collocation point in the operator domain and \(G_\theta\) is the DeepONet output. \(s^{(i)}(y_j^{(i)})\) is the initial loss represents the PDE solution at \(y_j^{(i)}\) conditioned on the \(i\)-th input function. For the Robin boundary condition \(\alpha\), \(\beta\), and \(\gamma\) are non-zero constants specified based on the physics of the problem. Similarly, \(L_{Physics}\) is defined as:

\begin{align}
\mathcal{L}_{physics}(\theta)  = & \frac{1}{NQm}\sum_{i=1}^{N}\sum_{j=1}^{Q}
\left|\mathcal{N}(u^{(i)}(x), G_{\theta}(u^{(i)})(y^{(i)}_j)) \right|^2
\end{align}

where \(\mathcal{N}\) is the nonlinear differential operator. In the above equations, \(N\) denotes the number of distinct input function combinations sampled from the design space and \(Q\) represents the number of residual points randomly sampled to enforce the physical constraints. They are considered hyperparameters and can be optimized based on the performance of the DeepONet and computational constraints. 

\subsection{Nonlinear decoder and multi-input functionality}

\begin{figure}[t]
\vskip 0.2in
\begin{center}
\centerline{\includegraphics[width=\columnwidth]{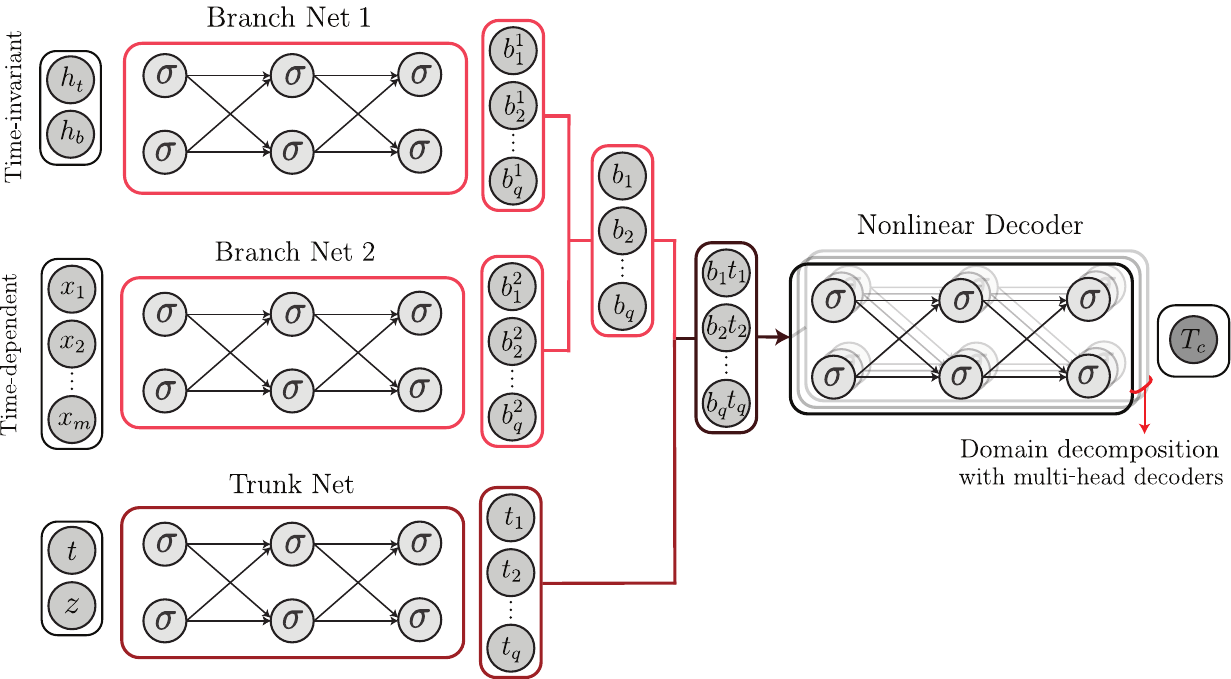}}
\caption{Schematic of proposed multi-input PIDON with NDs. Two branch nets are integrated to process time-dependent and time-independent input functions. NDs are responsible for learning the solution operators at different subdomains.}
\label{fig2}
\end{center}
\vskip -0.2in
\end{figure}

The original architecture of DeepONet uses a linear decoder to learn the nonlinear operator. This results in approximating the target functions with a finite-dimensional linear subspace. However, for cases where the target functional data is concentrated in nonlinear manifolds, the vanilla DeepONet can easily fail unless a very high-dimensional linear decoder architecture is implemented \cite{lanthaler2022error, lee2023hyperdeeponet}. This would result in a very large number of basis functions and coefficients determined by the output size of trunk and branch nets, respectively. Thus, the training of DeepONet can become computationally intensive for complex and non-smooth target functions. Different variants of DeepONet have been introduced to tackle the above limitation \cite{fang2024learning, haghighat2024deeponet}. One way to address this limitation is to integrate a nonlinear decoder (ND) into the architecture of DeepONet. \citet{seidman2022nomad} introduced NOMAD, a novel nonlinear manifold decoder leveraging neural networks to incorporate nonlinearity in the DeepONet’s decoder. In this approach, the model merges the input function sensor values with the query points, forming a concatenated input, which is then fed to the decoder’s network to predict the operator’s output. \citet{lee2023hyperdeeponet} proposed HyperDeepONet which substitutes the branch net with a hypernetwork to reduce the required network size for learning the solution operator. The hypernetwork is tasked with generating the weights of the trunk net given the input functions. Unlike the original architecture that distilled input function information into an embedding vector and fed it to the trunk net only in the final layer, the hypernetwork disperses this information at all layers of the trunk net. Here, while preserving the original architecture of DeepONet, we incorporate nonlinearity in the form of a fully-connected neural network immediately after merging the branch and trunk nets as depicted in Figure \ref{fig2}. Instead of summing the element-wise product of the trunk and branch networks’ last layer, we channel the resulting vector to a neural network responsible for capturing the nonlinearity in the target functions space. We show that replacing the linear layer with a neural network as the ND not only improves the performance of DeepONet in complex PDE systems but also allows learning nonlinear operators with low-dimensional feature representations in the output of branch and trunk nets.

The DeepONet architecture is originally designed to map a single input function to the target output function. This imposes constraints, particularly in the process design optimization of engineering systems where there's a necessity to optimize multiple input functions (i.e., design variables) simultaneously with the optimization of engineering systems' process designs. To overcome this limitation, the proposed architecture utilizes a multi-input functionality as depicted in Figure \ref{fig2}, which allows the neural operator to effectively process multiple input functions. In particular, one branch network is dedicated to processing the time-dependent process parameters (BN2) while the second branch net is tasked to collectively process all time-invariant design variables (BN1). This is as opposed to assigning a separate network to each process parameter \cite{kumar2023real}, which significantly reduces the computational cost during the training of the DeepONet. The output layer size of the branch nets is the same as that of the trunk net. The branch nets’ outputs \(b_1\) and \(b_2\) are merged via the Hadamard product, resulting in a \(q\)-dimensional vector \(b = \sum_{i=1}^{q}b_i^1b_i^2\). The resulting embedding which carries information about all input functions is then combined with the output of the trunk network, similar to vanilla DeepONet. 

\subsection{Domain decomposition}

Similar to PINN, PIDON also enforces governing laws by incorporating physical constraints in the form of additional loss terms and minimizing them using optimization algorithms such as gradient descent or its variants. Thus, naturally, PIDON would inherit the limitations PINN encountered when learning highly nonlinear and non-smooth systems \cite{wang2022and}. The presence of nonlinear time-varying characteristics and sharp transitions (e.g., stiff PDEs) hugely deteriorates the performance of physics-informed models. Long temporal domains and the F-principle effect in neural networks are other common reasons for the failure of physics-informed models \cite{wang2024respecting}. Specifically, the curing process of composites in the autoclave occurs over a long period of time and involves highly nonlinear characteristics. To improve the performance of PIDON in such a system, we decompose the PDE domain into smaller subdomains and learn each subdomain using a separate ND. This is similar to sequential learning \cite{mattey2022novel, wight2020solving} and extended PINN approaches proposed for the training of PINNs \cite{jagtap2020extended}. As illustrated in Figure \ref{fig2}, we implement a multi-head decoder architecture where the NDs share the same branch and trunk nets, with each subnet dedicated to learning a segment of the time domain. The size of the subdomains depends on the physical characteristics of the problem at hand. By splitting the temporal domain into smaller intervals, we effectively break down the problem into multiple PDE instances, easier to handle by the NDs. The time intervals can be spread uniformly across the domain or selected according to the complexity of the problem, e.g., more intervals are concentrated around sharp transitions and less where the behavior is not as chaotic and nonlinear. Formally, we decompose the domain of the PDE problem \(\Omega\) into \(N_d\) subdomains as \(\Omega =\bigcup_{k=1}^{N_d}\Omega_k\). Each subdomain \(\Omega_k\) is associated with an ND \(f_k(\theta_k)\) tasked to learn the solution of the PDE within its subdomain. The continuity between the subdomains is enforced via an additional \textit{interface} loss term, which is minimized along with the initial, boundary, and residual loss components during the training. The interface loss is calculated using the collocation points on the interface of adjacent subdomains \(\partial\Omega_p \cap \partial\Omega_q \) where \(i, j \in {1, 2, ..., N_d}\). The full solution of the PDE problem in the domain \(\Omega\) is achieved by combining all trained NDs. In this paper, the interface loss between the \(p\)-th and \(q\)-th subdomains is defined as:

\begin{align}
\mathcal{L}_{IF}(\theta_p, \theta_q) = & \frac{1}{NQ_{if}} \sum_{i=1}^{N} \sum_{j=1}^{Q_{if}} 
\left| G_{\theta_p}(u^{(i)})(y^{(i)}_j) \right. \notag \\
& \left. - G_{\theta_q}(u^{(i)})(y^{(i)}_j) \right|^2
\end{align}

\subsection{Decoupled DeepONets for multi-output prediction} \label{decoupledsec}

\begin{figure*}[t]
\vskip 0.2in
\begin{center}
\centerline{\includegraphics[height=5cm]{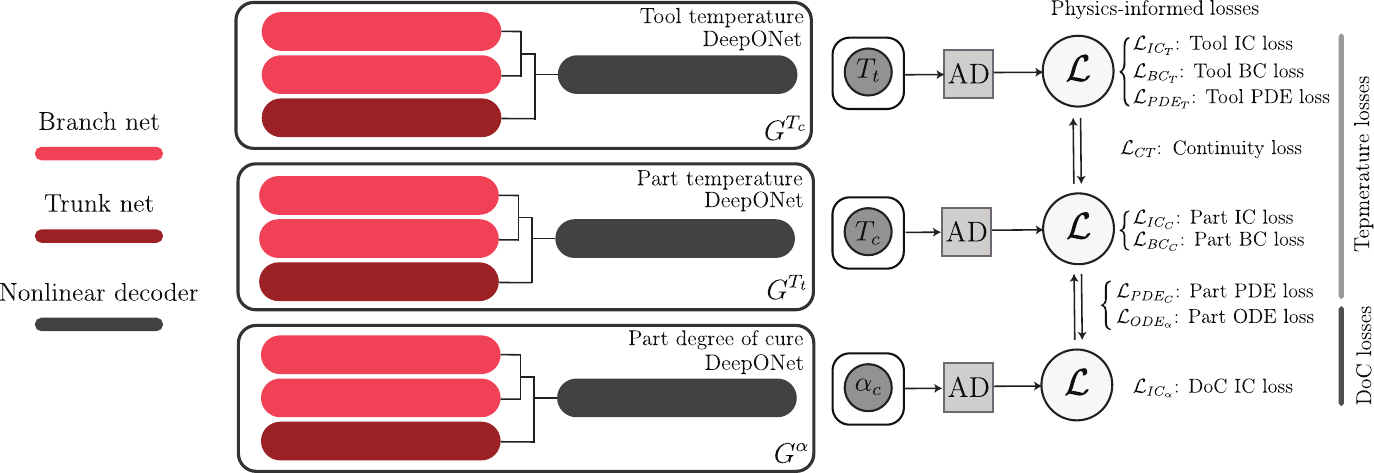}}
\caption{Schematic of proposed multi-input PIDON with nonlinear decoder for thermochemical analysis of composites curing process.}
\label{EnPIDON}
\end{center}
\vskip -0.2in
\end{figure*}

As elaborated in Section \ref{compositessection}, the thermochemical analysis of composites during the curing process requires learning multiple target functions, namely, part temperature, tooling temperature, and part DoC. One way to achieve this is to devise multiple neurons in the output layer of DeepONet’s ND, where each neuron is responsible for predicting one of the output functions. However, we observed that this configuration results in poor performance, mainly due to the significant discrepancies in the behavior of output functions and the limitation of the shared DeepONet architecture to capture those fully. Another approach would be to introduce a multi-head functionality where each output function has its own dedicated ND. The decoders take the branch and trunk networks’ embedding as the shared input and separately learn the mappings to each target function. However, while more effective than the first strategy, it is yet unable to accurately learn the system’s solution. We hypothesize that the presence of distinct thermal behaviors in this bi-material system as well as its multi-scale physics poses a challenge for a single branch-trunk network to effectively support the learning of the entire temperature and DoC fields. To address this, we utilize a fully decoupled DeepONet architecture where each output function has its own dedicated branch net, trunk net, and ND, as shown in Figure \ref{EnPIDON}. This design yields three distinct neural operators denoted as \(G^{T_c}\), \(G^{T_t}\), and \(G^{\alpha}\). It is worth noting that decoupling the output functions automatically imposes a spatial domain decomposition as the temperature profiles of the part and tooling are learned via 2 separate DeepONets. This essentially allows some of NDs to concentrate exclusively on understanding the tooling’s thermal characteristics, while the rest are dedicated to learning the thermochemical behavior of the composite part.

\subsection{Curriculum learning}

It has been shown that the vanilla PINN model has difficulty learning highly nonlinear PDEs with large coefficients \cite{krishnapriyan2021characterizing}. This can be due to the sensitivity of the PINN's loss landscape to the values of these coefficients. Particularly, when coefficients have large values, the loss landscape tends to become complex and asymmetric, posing significant challenges to the training process. To mitigate this issue, one effective approach involves utilizing curriculum learning strategies \cite{krishnapriyan2021characterizing, bengio2009curriculum}. By initially training the model on PDE solutions with smaller coefficients (i.e., smoother loss landscape) and gradually increasing the coefficient values, the model error can be reduced significantly by several orders of magnitude. We also observed similar challenges in training the PIDON model. Specifically, in the context of thermochemical analysis of composite curing processes, the presence of a heat generation term within the heat transfer governing equation introduces a sharp nonlinearity into the PDE solution (\ref{CureEquations}). This nonlinearity substantially contributes to the poor predictive performance of physics-informed models. Particularly for large values of the heat generation coefficient (\(b_c\)), training the PIDON model becomes notably challenging. To address this issue, we employed the curriculum learning strategy described above. This involves initiating the training process with no heat generation term (\(b_c = 0\)) and gradually introducing internal heat generation to the equation through step-wise increments in the value of \(b_c\).

\section{Results} \label{results}

This section presents a series of experimental results demonstrating the effectiveness of the nonlinear decoders, curriculum learning and domain decomposition in the training of PIDON. The performance of PIDON against the highly nonlinear composites curing process across a high-dimensional design space is investigated. For model training and evaluation, \(500\) and \(20\) random combinations of input functions were generated from the specified ranges presented in Table \ref{designRange}. All branch, trunk and NDs of DeepONet models consist of 5 hidden layers with 50 neurons equipped with tanh activation function. A 50-neuron output layer is selected for both branch and trunk nets. PIDON was trained using Adam optimizer with an initial learning rate of \(1 \times  10^{-3}\) and a decay rate of \(0.9\) per \(1000\) steps. A batch size of \(1024\) and \(200\) training epochs was employed. The Jax library \cite{jax2018github} was used for developing and training the models on a single NVIDIA T4 GPU with 104 GB of memory. For validation, an in-house Python FE code was developed and used to randomly generate unseen test cases from the design spaces. AS4/8552 prepreg and Invar tooling are considered as the materials for this case study. The model's average performance on these unseen test cases is reported. Details regarding the training procedure of PIDON models are presented in \ref{Trainingprocess}.

\subsection{Evaluation of PIDON's predictive performance}

We trained the PIDON model on the design space characterized by the design variables outlined in Section \ref{compositessection}. Three DeepONets equipped with NDs were trained to predict the output functions, specifically \(T_c\), \(T_t\), and \(\alpha\). The training of DeepONets was conducted sequentially \cite{niaki2021physics}, where two Adam optimizers sequentially minimize temperature- and DoC-related losses (Figure \ref{EnPIDON}) for improved stability and convergence. To capture the complex dynamics of the composite part, the time domain was decomposed into 7 intervals, with smaller intervals centered around the DoC sharp transition (Figure \ref{pidon_mapping}). The parametric coupled PDEs were learned via a curriculum learning strategy by incrementally increasing the heat generation coefficient \(b_c\) from 0 (no heat generation) to its real value in 5 steps.

\begin{figure}[!t]
\vskip 0.2in
\begin{center}
\centerline{\includegraphics[width=\columnwidth]{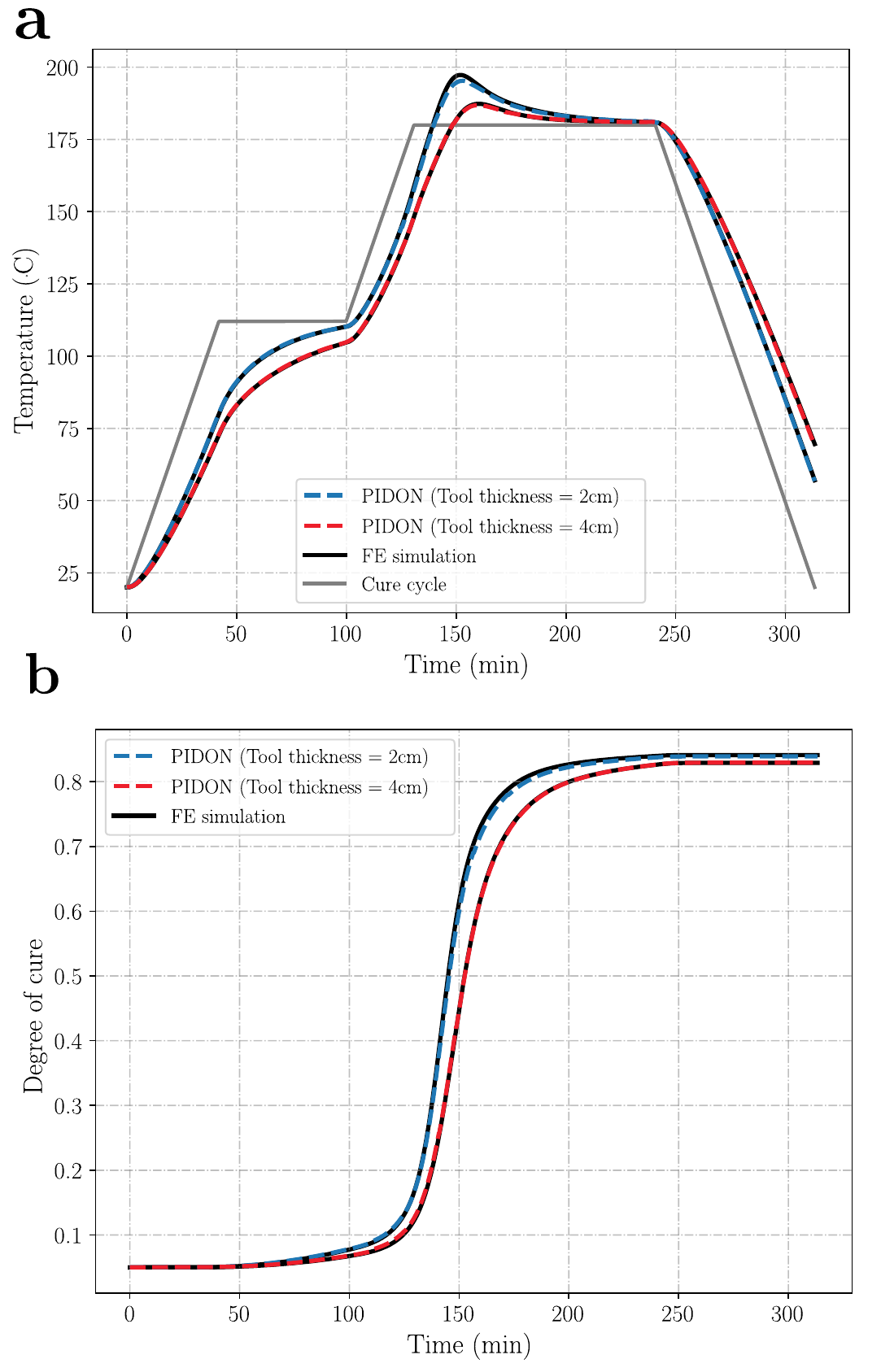}}
\caption{Temperature (a) and DoC (b) prediction performance of PIDON at composite part mid-point for two curing scenarios: thin tooling (\(T\) maximum absolute error = 3.2°C, \(\alpha\) maximum error = 0.024) and thick tooling (\(T\) maximum error = 0.9°C, \(\alpha\) maximum absolute error = 0.021). An identical two-hold cure cycle was used for both scenarios (shown in gray.)}
\label{tool_comp}
\end{center}
\vskip -0.2in
\end{figure}

Figures \ref{tool_comp} and \ref{morepred} illustrates the model's predictions of the part's mid-point temperature and DoC across various design variable combinations. The model achieved an average maximum absolute error of 2.3°C and 0.022 for temperature and DoC across test cases.  Figure \ref{imshow} provides a visual comparison between PIDON's predictions and FE simulations, along with the absolute error fields for part temperature, part DoC, and Tool temperature. The real-time inference capability of PIDON (20 times faster than FE simulations in this case) as well as its accurate predictions across a high-dimensional design space, makes it an excellent tool for process design optimization tasks.

\subsection{Effect of nonlinear decoder}
\begin{figure}[!t]
\vskip 0.2in
\begin{center}
\centerline{\includegraphics[width=\columnwidth]{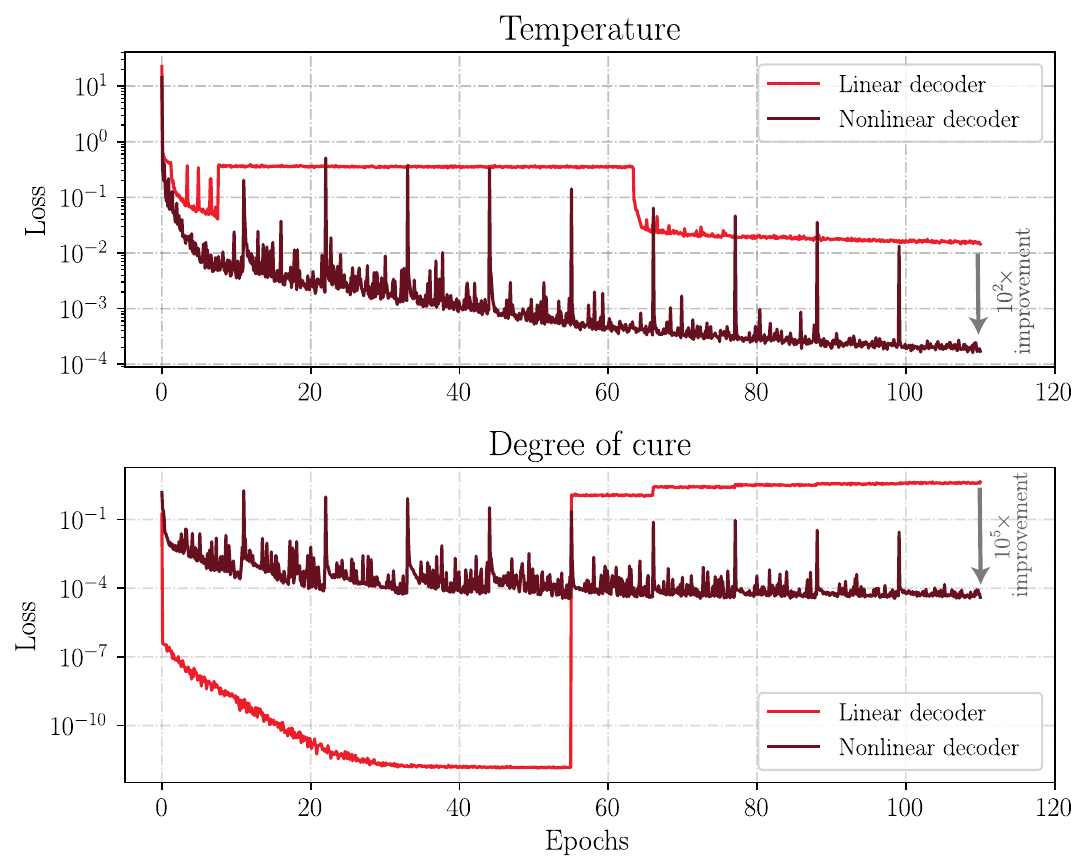}}
\caption{Effect of implementing ND in the architecture of DeepONet. The model with ND (dark red) results in considerably smaller training loss and exhibits a more stable behavior.}
\label{fig4}
\end{center}
\vskip -0.2in
\end{figure}

To showcase the impact of integrating NDs into the architecture of PIDON, two training scenarios were considered: one with NDs and one without (using the original DeepONet's linear decoder.) As depicted in Figure \ref{fig4}, the model lacking the ND struggles to capture the complexities within the solution, ultimately converging to a relatively high training loss. Despite experimenting with various sizes of branch and trunk output layers within the vanilla DeepONet architecture, the model's performance remains unsatisfactory. In contrast, the PIDON model equipped with ND successfully learns the underlying physics, resulting in a significantly reduced training loss. Moreover, the addition of ND enables a more stable training process for learning the solution of the coupled PDEs, concurrently reducing both temperature and DoC losses to satisfactory levels. Conversely, PIDON's linear decoder initially converges to a trivial solution by equating the rate of change of DoC to zero (hence, a very small DoC loss at the initial stage of training). This, consequently, prevents the temperature loss from decreasing to small values. Upon exiting the trivial solution, while the temperature loss is improved, the DoC loss increases significantly and remains at large values. This observation underscores the inability of the linear decoder in the original architecture of DeepONet to learn nonlinear operators. 

\subsection{Curriculum learning}

While utilizing NDs significantly improves the performance of PIDON, we still observed notable deviation between the PIDON’s predictions and the FE solutions, particularly around sharp boundary edges and nonlinear trends during the steep rise of DoC. This discrepancy is exacerbated as the problem complexity increases with broadening the input function range and expanding the design space. We hypothesize that one contributing factor is the nonsmooth, asymmetric, and complex loss landscape of the PIDON, hindering convergence to small loss values \cite{krishnapriyan2021characterizing}. The presence of multiple loss components, including PDE, ODE, IC, BC, interface, and continuity losses compounds the optimization challenge. Here, we explored the impact of a curriculum learning strategy across three different design space sizes. Specifically, PIDONs were trained with and without curriculum learning on small, medium, and large input function ranges. For curriculum learning, training commenced with solving a simplified heat transfer problem without heat generation by setting the heat generation coefficient \(b_c\) (\ref{CureEquations}) to zero. Then, the training gradually progressed to more complex scenarios by incrementally increasing the value of \(b_c\) in a step-wise fashion (Figure \ref{CLprogress}). At each stage, the trained weights from the previous step served as initialization. Table \ref{CL} presents a comparative analysis of PIDON performance with and without curriculum learning across various design space sizes. While PIDON accuracy diminishes with larger design spaces, those trained with curriculum learning consistently exhibit strong performance across all design space scales.

\begin{table}[t]
\caption{Comparison of PIDON's temperature prediction performance with and without curriculum learning across different design space sizes. The description of design spaces is provided in \ref{Trainingprocess}. All models were trained using 7 NDs.}
\label{CL}
\vskip 0.15in
\begin{center}
\begin{small}
\begin{sc}
\begin{tabular}{lcccr}
\toprule
Design     & Metric & Regular  & Curriculum  \\
space size &   (Avg.)     & training & learning \\
\midrule
Small    & Rel. $L^2$   & $6.8\times10^{-3}$   & $2.8\times10^{-3}$\\
         & MAE          & 0.74    & 0.285    \\[.15cm] 

Medium   & Rel. $L^2$   & $9.22\times10^{-3}$    & $3.27\times10^{-3}$   \\
         & MAE          & 0.83    & 0.361    \\[.15cm] 

Large    & Rel. $L^2$   & $1.3\times10^{-2}$    & $4.35\times10^{-3}$   \\
         & MAE          & 1.1    & 0.447    \\

\bottomrule
\end{tabular}
\footnotetext{random text}
\end{sc}
\end{small}
\end{center}
\vskip -0.1in
\end{table}

\subsection{Role of domain decomposition in capturing nonlinearities} \label{EffectDD}

We conducted experiments to evaluate the efficacy of domain decomposition in enhancing the performance of PIDON for modeling the thermochemical analysis of composites curing process. Notably,  while curriculum learning improved the model’s overal performance, we observed that it still struggled to fully capture the behavior around sharp nonlinearities, particularly during the rapid increase in the DoC, responsible for the heat generation phenomena. This directly affects the model’s performance on the prediction of exotherm (maximum part temperautre), a key factor in determining the quality of the manufactured part. To address this limitation, we utilized domain decomposition by partitioning the time domain into smaller intervals centered on the nonlinear region and allocating NDs to learn the physics within each interval. This approach further improved the model's performance around the nonlinearities. Specifically, we tested three architectures with different number of subnets as shown in Table \ref{DD}. All models followed an identical training procedure, including the implementation of the curriculum learning strategy and an equal number of training epochs. The exotherm prediction error decreased substantially as the number of NDs increased. These results suggest that despite the effectiveness of curriculum learning in mitigating the adverse impacts of nonsmooth landscapes associated with large PDE coefficients, achieving more accurate predictions in highly nonlinear regimes necessitates more expressivity, which can be attained through the introduction of multiple NDs.

\begin{table}[!t]
\caption{Effect of Domain Decomposition on PIDON's temperature prediction performance in highly nonlinear regions.}
\label{DD}
\vskip 0.15in
\begin{center}
\begin{small}
\begin{sc}
\begin{tabular}{lccc}
\toprule
Metric & \multicolumn{3}{c}{Number of NDs ($N_d$)} \\
  (Avg.)       &  1       & 5           & 7      \\
\midrule
$\textup{Rel.} \  L^2 (\times10^{-3})$    & $6.1$& $3.6$ & $2.8$  \\
$\textup{Max error (°C)}$ & 6.1 & 3.1& 2.3\\
$\textup{Training time (s/epc.)}$ & 40 &56 & 61      \\

\bottomrule
\end{tabular}
\end{sc}
\end{small}
\end{center}
\vskip -0.1in
\end{table}

\section*{Conclusions} \label{conclusion}
In this study, we presented an advanced PIDON framework designed to address the challenges posed by highly nonlinear and complex physical systems, specifically in the context of composites autoclave processing. Our approach integrates nonlinear decoders, domain decomposition, and curriculum learning strategies, which together notably improve the model’s ability to capture complex solution operators and provide accurate predictions and generalizability across a wide range of input functions and design spaces. The introduced enhancements effectively addressed the shortcomings of vanilla PIDON architecture, resulting in a robust and reliable predictive performance. The advanced PIDON's zero-shot and real time inference capabilities make it highly suitable for applications in digital twins and Industry 4.0, where real-time data and simulations are crucial for monitoring and controlling manufacturing processes. This ability to provide swift and precise insights ensures that PIDON can play a pivotal role in enhancing the efficiency and reliability of composites manufacturing and other related fields.


\bibliography{example_paper}

\begin{thebibliography}{41}
\providecommand{\natexlab}[1]{#1}
\providecommand{\url}[1]{\texttt{#1}}
\expandafter\ifx\csname urlstyle\endcsname\relax
  \providecommand{\doi}[1]{doi: #1}\else
  \providecommand{\doi}{doi: \begingroup \urlstyle{rm}\Url}\fi

\bibitem[Bengio et~al.(2009)Bengio, Louradour, Collobert, and Weston]{bengio2009curriculum}
Bengio, Y., Louradour, J., Collobert, R., and Weston, J.
\newblock Curriculum learning.
\newblock In \emph{Proceedings of the 26th annual international conference on machine learning}, pp.\  41--48, 2009.

\bibitem[Boull{\'e} \& Townsend(2023)Boull{\'e} and Townsend]{boulle2023mathematical}
Boull{\'e}, N. and Townsend, A.
\newblock A mathematical guide to operator learning.
\newblock \emph{arXiv preprint arXiv:2312.14688}, 2023.

\bibitem[Bradbury et~al.(2018)Bradbury, Frostig, Hawkins, Johnson, Leary, Maclaurin, Necula, Paszke, Vander{P}las, Wanderman-{M}ilne, and Zhang]{jax2018github}
Bradbury, J., Frostig, R., Hawkins, P., Johnson, M.~J., Leary, C., Maclaurin, D., Necula, G., Paszke, A., Vander{P}las, J., Wanderman-{M}ilne, S., and Zhang, Q.
\newblock {JAX}: composable transformations of {P}ython+{N}um{P}y programs, 2018.
\newblock URL \url{http://github.com/google/jax}.

\bibitem[Chen et~al.(2021)Chen, Li, Meng, Zhou, Hao, et~al.]{chen2021residual}
Chen, G., Li, Y., Meng, Q., Zhou, J., Hao, X., et~al.
\newblock Residual fourier neural operator for thermochemical curing of composites.
\newblock \emph{arXiv preprint arXiv:2111.10262}, 2021.

\bibitem[Chen et~al.(2023)Chen, Li, Liu, Mehdi-Souzani, Meng, Zhou, and Hao]{chen2023physics}
Chen, G., Li, Y., Liu, X., Mehdi-Souzani, C., Meng, Q., Zhou, J., and Hao, X.
\newblock Physics-guided neural operator for data-driven composites manufacturing process modelling.
\newblock \emph{Journal of Manufacturing Systems}, 70:\penalty0 217--229, 2023.

\bibitem[Chen \& Chen(1995)Chen and Chen]{chen1995universal}
Chen, T. and Chen, H.
\newblock Universal approximation to nonlinear operators by neural networks with arbitrary activation functions and its application to dynamical systems.
\newblock \emph{IEEE transactions on neural networks}, 6\penalty0 (4):\penalty0 911--917, 1995.

\bibitem[Cuomo et~al.(2022)Cuomo, Di~Cola, Giampaolo, Rozza, Raissi, and Piccialli]{cuomo2022scientific}
Cuomo, S., Di~Cola, V.~S., Giampaolo, F., Rozza, G., Raissi, M., and Piccialli, F.
\newblock Scientific machine learning through physics--informed neural networks: Where we are and what’s next.
\newblock \emph{Journal of Scientific Computing}, 92\penalty0 (3):\penalty0 88, 2022.

\bibitem[Fabris(2018)]{fabris2018framework}
Fabris, J.~N.
\newblock \emph{A framework for formalizing science based composites manufacturing practice}.
\newblock PhD thesis, University of British Columbia, 2018.

\bibitem[Fang et~al.(2024)Fang, Wang, and Perdikaris]{fang2024learning}
Fang, Z., Wang, S., and Perdikaris, P.
\newblock Learning only on boundaries: A physics-informed neural operator for solving parametric partial differential equations in complex geometries.
\newblock \emph{Neural Computation}, 36\penalty0 (3):\penalty0 475--498, 2024.

\bibitem[Gao et~al.(2021)Gao, Sun, and Wang]{gao2021phygeonet}
Gao, H., Sun, L., and Wang, J.-X.
\newblock Phygeonet: Physics-informed geometry-adaptive convolutional neural networks for solving parameterized steady-state pdes on irregular domain.
\newblock \emph{Journal of Computational Physics}, 428:\penalty0 110079, 2021.

\bibitem[Gin et~al.(2021)Gin, Shea, Brunton, and Kutz]{gin2021deepgreen}
Gin, C.~R., Shea, D.~E., Brunton, S.~L., and Kutz, J.~N.
\newblock Deepgreen: deep learning of green’s functions for nonlinear boundary value problems.
\newblock \emph{Scientific reports}, 11\penalty0 (1):\penalty0 21614, 2021.

\bibitem[Haghighat et~al.(2024)Haghighat, bin Waheed, and Karniadakis]{haghighat2024deeponet}
Haghighat, E., bin Waheed, U., and Karniadakis, G.
\newblock En-deeponet: An enrichment approach for enhancing the expressivity of neural operators with applications to seismology.
\newblock \emph{Computer Methods in Applied Mechanics and Engineering}, 420:\penalty0 116681, 2024.

\bibitem[Hubert et~al.(2001)Hubert, Johnston, Poursartip, and Nelson]{hubert2001cure}
Hubert, P., Johnston, A., Poursartip, A., and Nelson, K.
\newblock Cure kinetics and viscosity models for hexcel 8552 epoxy resin.
\newblock In \emph{International SAMPE symposium and exhibition}, pp.\  2341--2354. SAMPE; 1999, 2001.

\bibitem[Jagtap \& Karniadakis(2020)Jagtap and Karniadakis]{jagtap2020extended}
Jagtap, A.~D. and Karniadakis, G.~E.
\newblock Extended physics-informed neural networks (xpinns): A generalized space-time domain decomposition based deep learning framework for nonlinear partial differential equations.
\newblock \emph{Communications in Computational Physics}, 28\penalty0 (5), 2020.

\bibitem[Krishnapriyan et~al.(2021)Krishnapriyan, Gholami, Zhe, Kirby, and Mahoney]{krishnapriyan2021characterizing}
Krishnapriyan, A., Gholami, A., Zhe, S., Kirby, R., and Mahoney, M.~W.
\newblock Characterizing possible failure modes in physics-informed neural networks.
\newblock \emph{Advances in Neural Information Processing Systems}, 34:\penalty0 26548--26560, 2021.

\bibitem[Kumar et~al.(2023)Kumar, Goswami, Smith, and Karniadakis]{kumar2023real}
Kumar, V., Goswami, S., Smith, D.~J., and Karniadakis, G.~E.
\newblock Real-time prediction of gas flow dynamics in diesel engines using a deep neural operator framework.
\newblock \emph{arXiv preprint arXiv:2304.00567}, 2023.

\bibitem[Lanthaler et~al.(2022)Lanthaler, Mishra, and Karniadakis]{lanthaler2022error}
Lanthaler, S., Mishra, S., and Karniadakis, G.~E.
\newblock Error estimates for deeponets: A deep learning framework in infinite dimensions.
\newblock \emph{Transactions of Mathematics and Its Applications}, 6\penalty0 (1):\penalty0 tnac001, 2022.

\bibitem[Lee et~al.(2023)Lee, Cho, and Hwang]{lee2023hyperdeeponet}
Lee, J.~Y., Cho, S.~W., and Hwang, H.~J.
\newblock Hyperdeeponet: learning operator with complex target function space using the limited resources via hypernetwork.
\newblock \emph{arXiv preprint arXiv:2312.15949}, 2023.

\bibitem[Li et~al.(2020{\natexlab{a}})Li, Kovachki, Azizzadenesheli, Liu, Bhattacharya, Stuart, and Anandkumar]{li2020fourier}
Li, Z., Kovachki, N., Azizzadenesheli, K., Liu, B., Bhattacharya, K., Stuart, A., and Anandkumar, A.
\newblock Fourier neural operator for parametric partial differential equations.
\newblock \emph{arXiv preprint arXiv:2010.08895}, 2020{\natexlab{a}}.

\bibitem[Li et~al.(2020{\natexlab{b}})Li, Kovachki, Azizzadenesheli, Liu, Bhattacharya, Stuart, and Anandkumar]{li2020neural}
Li, Z., Kovachki, N., Azizzadenesheli, K., Liu, B., Bhattacharya, K., Stuart, A., and Anandkumar, A.
\newblock Neural operator: Graph kernel network for partial differential equations.
\newblock \emph{arXiv preprint arXiv:2003.03485}, 2020{\natexlab{b}}.

\bibitem[Lu et~al.(2021)Lu, Jin, Pang, Zhang, and Karniadakis]{lu2021learning}
Lu, L., Jin, P., Pang, G., Zhang, Z., and Karniadakis, G.~E.
\newblock Learning nonlinear operators via deeponet based on the universal approximation theorem of operators.
\newblock \emph{Nature machine intelligence}, 3\penalty0 (3):\penalty0 218--229, 2021.

\bibitem[Lu et~al.(2022)Lu, Meng, Cai, Mao, Goswami, Zhang, and Karniadakis]{lu2022comprehensive}
Lu, L., Meng, X., Cai, S., Mao, Z., Goswami, S., Zhang, Z., and Karniadakis, G.~E.
\newblock A comprehensive and fair comparison of two neural operators (with practical extensions) based on fair data.
\newblock \emph{Computer Methods in Applied Mechanics and Engineering}, 393:\penalty0 114778, 2022.

\bibitem[Lucia et~al.(2004)Lucia, Beran, and Silva]{lucia2004reduced}
Lucia, D.~J., Beran, P.~S., and Silva, W.~A.
\newblock Reduced-order modeling: new approaches for computational physics.
\newblock \emph{Progress in aerospace sciences}, 40\penalty0 (1-2):\penalty0 51--117, 2004.

\bibitem[Majumdar et~al.(2024)Majumdar, Jadhav, Deodhar, Karande, Vig, and Runkana]{majumdar2024hxpinn}
Majumdar, R., Jadhav, V., Deodhar, A., Karande, S., Vig, L., and Runkana, V.
\newblock Hxpinn: A hypernetwork-based physics-informed neural network for real-time monitoring of an industrial heat exchanger.
\newblock \emph{Numerical Heat Transfer, Part B: Fundamentals}, pp.\  1--22, 2024.

\bibitem[Mattey \& Ghosh(2022)Mattey and Ghosh]{mattey2022novel}
Mattey, R. and Ghosh, S.
\newblock A novel sequential method to train physics informed neural networks for allen cahn and cahn hilliard equations.
\newblock \emph{Computer Methods in Applied Mechanics and Engineering}, 390:\penalty0 114474, 2022.

\bibitem[Meng et~al.(2023)Meng, Li, Liu, Chen, and Hao]{meng2023novel}
Meng, Q., Li, Y., Liu, X., Chen, G., and Hao, X.
\newblock A novel physics-informed neural operator for thermochemical curing analysis of carbon-fibre-reinforced thermosetting composites.
\newblock \emph{Composite Structures}, 321:\penalty0 117197, 2023.

\bibitem[Niaki et~al.(2021)Niaki, Haghighat, Campbell, Poursartip, and Vaziri]{niaki2021physics}
Niaki, S.~A., Haghighat, E., Campbell, T., Poursartip, A., and Vaziri, R.
\newblock Physics-informed neural network for modelling the thermochemical curing process of composite-tool systems during manufacture.
\newblock \emph{Computer Methods in Applied Mechanics and Engineering}, 384:\penalty0 113959, 2021.

\bibitem[Rahman et~al.(2024)Rahman, George, Elleithy, Leibovici, Li, Bonev, White, Berner, Yeh, Kossaifi, et~al.]{rahman2024pretraining}
Rahman, M.~A., George, R.~J., Elleithy, M., Leibovici, D., Li, Z., Bonev, B., White, C., Berner, J., Yeh, R.~A., Kossaifi, J., et~al.
\newblock Pretraining codomain attention neural operators for solving multiphysics pdes.
\newblock \emph{arXiv preprint arXiv:2403.12553}, 2024.

\bibitem[Raissi et~al.(2019)Raissi, Perdikaris, and Karniadakis]{raissi2019physics}
Raissi, M., Perdikaris, P., and Karniadakis, G.~E.
\newblock Physics-informed neural networks: A deep learning framework for solving forward and inverse problems involving nonlinear partial differential equations.
\newblock \emph{Journal of Computational physics}, 378:\penalty0 686--707, 2019.

\bibitem[Ramezankhani \& Milani(2023)Ramezankhani and Milani]{ramezankhani2023sequential}
Ramezankhani, M. and Milani, A.~S.
\newblock A sequential meta-transfer (smt) learning to combat complexities of physics-informed neural networks: Application to composites autoclave processing.
\newblock \emph{arXiv preprint arXiv:2308.06447}, 2023.

\bibitem[Rashid et~al.(2022)Rashid, Pittie, Chakraborty, and Krishnan]{rashid2022learning}
Rashid, M.~M., Pittie, T., Chakraborty, S., and Krishnan, N.~A.
\newblock Learning the stress-strain fields in digital composites using fourier neural operator.
\newblock \emph{Iscience}, 25\penalty0 (11), 2022.

\bibitem[Seidman et~al.(2022)Seidman, Kissas, Perdikaris, and Pappas]{seidman2022nomad}
Seidman, J., Kissas, G., Perdikaris, P., and Pappas, G.~J.
\newblock Nomad: Nonlinear manifold decoders for operator learning.
\newblock \emph{Advances in Neural Information Processing Systems}, 35:\penalty0 5601--5613, 2022.

\bibitem[Sirignano \& Spiliopoulos(2018)Sirignano and Spiliopoulos]{sirignano2018dgm}
Sirignano, J. and Spiliopoulos, K.
\newblock Dgm: A deep learning algorithm for solving partial differential equations.
\newblock \emph{Journal of computational physics}, 375:\penalty0 1339--1364, 2018.

\bibitem[Strong(2008)]{strong2008fundamentals}
Strong, A.~B.
\newblock \emph{Fundamentals of composites manufacturing: materials, methods and applications}.
\newblock Society of manufacturing engineers, 2008.

\bibitem[Tripura \& Chakraborty(2022)Tripura and Chakraborty]{tripura2022wavelet}
Tripura, T. and Chakraborty, S.
\newblock Wavelet neural operator: a neural operator for parametric partial differential equations.
\newblock \emph{arXiv preprint arXiv:2205.02191}, 2022.

\bibitem[Van~Ee \& Poursartip(2009)Van~Ee and Poursartip]{van2009hexply}
Van~Ee, D. and Poursartip, A.
\newblock Hexply 8552 material properties database for use with compro cca and raven.
\newblock \emph{Version 0.9. NCAMP. Wichita, KS}, 2009.

\bibitem[Wang et~al.(2021)Wang, Wang, and Perdikaris]{wang2021learning}
Wang, S., Wang, H., and Perdikaris, P.
\newblock Learning the solution operator of parametric partial differential equations with physics-informed deeponets.
\newblock \emph{Science advances}, 7\penalty0 (40):\penalty0 eabi8605, 2021.

\bibitem[Wang et~al.(2022)Wang, Yu, and Perdikaris]{wang2022and}
Wang, S., Yu, X., and Perdikaris, P.
\newblock When and why pinns fail to train: A neural tangent kernel perspective.
\newblock \emph{Journal of Computational Physics}, 449:\penalty0 110768, 2022.

\bibitem[Wang et~al.(2024)Wang, Sankaran, and Perdikaris]{wang2024respecting}
Wang, S., Sankaran, S., and Perdikaris, P.
\newblock Respecting causality for training physics-informed neural networks.
\newblock \emph{Computer Methods in Applied Mechanics and Engineering}, 421:\penalty0 116813, 2024.

\bibitem[Wight \& Zhao(2020)Wight and Zhao]{wight2020solving}
Wight, C.~L. and Zhao, J.
\newblock Solving allen-cahn and cahn-hilliard equations using the adaptive physics informed neural networks.
\newblock \emph{arXiv preprint arXiv:2007.04542}, 2020.

\bibitem[Zauderer(2011)]{zauderer2011partial}
Zauderer, E.
\newblock \emph{Partial differential equations of applied mathematics}.
\newblock John Wiley \& Sons, 2011.

\end{thebibliography}
\bibliographystyle{icml2024}

\newpage
\appendix
\onecolumn
\section{Composites autoclave processing case study}

\subsection{Governing equations} \label{CureEquations}
The one-dimensional thermochemical behavior of a composite-tool system in an autoclave is governed by an anisotropic heat conduction equation with an internal heat generation term \(\dot{Q}=b_c\frac{\partial \alpha}{\partial t}\) accounting for the exothermic chemical reaction of the resin matrix during the curing process:

\begin{equation}
\left\{ \begin{aligned} 
  \frac{\partial T_t}{\partial t} &= a_t\frac{\partial^2T_t}{\partial z^2} \ \ \ \ \ \ \ \ \ \ \ \ \ \ \ \  \ \ \  z\in [0, L_1]
  \\
  \frac{\partial T_c}{\partial t} &= a_c\frac{\partial^2T_c}{\partial z^2}+ b_c\frac{\partial \alpha}{\partial t} \ \ \ \ \  z\in [L_1, L_2]
\end{aligned} \right. \  \ \  \ 
\textup{where} \ \  a = \frac{k}{\rho C_p} \ \ \textup{and} \ \ b = \frac{v_r\rho_rH_r}{\rho C_p}.
\end{equation}

where, \(T\) is the temperature \(\alpha\) is the DoC, \(L\) is the material length, and \(t\) and \(z\) are the spatiotemporal coordinates. Subscripts \(t\), \(c\), and \(r\), represent the tool, composite part and resin, respectively. \(a \) denotes the thermal diffusivity, \(b\) is the heat generation coefficient, and \(k\), \(p\) and \(C_p\) are the thermal conductivity, density and specific heat capacity. \(v\) and \(H\) represent the volume fraction and heat of reaction per unit mass. In the curing process of a composite system with thermoset resin, the cure rate \( \frac{\partial \alpha}{\partial t}\) is determined by the resin's cure kinetics and is typically described by an ordinary differential equation. For the 8552 epoxy resin system, used in this study, the cure kinetics have been previously developed \cite{hubert2001cure}, and can be expressed as follows:

\begin{equation}
    \frac{\partial \alpha}{\partial t}=\frac{A\textup{ exp}(-\frac{\Delta E}{RT}))}{1+\textup{exp}(C(\alpha - (C_0 + C_TT)))}\alpha^m(1-\alpha)^n.
\end{equation}

Here, \(\Delta E\) represents the activation energy, \(R\) is the gas constant, and \(C_0\), \(C_T\), \(m\), \(n\) and \(A\) are experimentally determined constants. Table \ref{kinetics} provides a summary of the parameter values used in the cure kinetics equations for this study. 
\begin{table}[h]
\caption{Summary of parameters used in heat transfer and cure kinetics governing equations.}
\label{kinetics}
\vskip 0.15in
\begin{center}
\begin{small}
\begin{sc}
\begin{tabular}{lllll}
    \toprule
    Parameter   & Description   & Value         \\
    \midrule
    \( \Delta E\) & Activation energy  & 66.5 \(\textup{(kJ/gmol)}\)    \\
    \( R\)     & Gas constant & 8.314     \\
    \( A\)     & Pre-exponential cure rate coefficient      & $ 1.53 \times 10^5 $ (1/s)  \\
    \( m\)     & First exponential constant      & 0.813  \\
    \( n\)     & Second exponential constant       & 2.74  \\
    \( C\)     & Diffusion constant       & 43.1  \\
    \( C_0\)     & Critical degree of cure at $T=0 $ K       & -1.684  \\
    \( C_T\)     & Critical resin degree of cure constant       & $5.475 \times 10^{-3} $ (1/K)  \\

    \bottomrule
\end{tabular}
\end{sc}
\end{small}
\end{center}
\vskip -0.1in
\end{table}

The initial conditions of the coupled system described above can be specified as:

\begin{equation}
\begin{aligned}
T_c\mid _{t=0} = T_0(x) \\
T_t\mid _{t=0} = T_0(x) \\
\alpha\mid _{t=0} = \alpha_0(x) .
\end{aligned}
\end{equation}

\(T_0\) represents the part's initial temperature, which is typically considered uniform throughout. In this study, the initial temperature is assumed to be 20°C. \(\alpha_0\) denotes the initial DoC of the resin system, and for an uncured part, it is assumed to be zero or a very small value; in this study, a value of 0.05 is used. Considering the convective heat transfer between the autoclave air \(T_a\) and the composites system, the boundary conditions are governed by:

\begin{equation}
\begin{aligned}
(T_a-T_c\mid _{z=L_2}) &= \frac{k_c}{h_{top}}\frac{\partial T_c}{\partial z}\mid _{z=L_2} \\
(T_t\mid _{z=0} - T_a) &= \frac{k_t}{h_{bot}}\frac{\partial T_t}{\partial z}\mid _{z=0}
\end{aligned}
\end{equation}

where \(h_{top}\) and \(h_{bot}\) are the HTCs on the top and bottom surfaces of the composite-tool system. Furthermore, the solution of the described system must satisfy the following continuity conditions between the part and the tool:

\begin{equation}
\begin{aligned}
k_t\frac{\partial T_t}{\partial {z}}\mid _{z=L_1^-} &= k_c\frac{\partial T_c}{\partial {z}}\mid _{z=L_1^+}
\\
T_t\mid _{z=L_1^-} &= T_c\mid _{z=L_1^+}.
\end{aligned}
\end{equation}

\begin{figure}[h]
\vskip 0.2in
\begin{center}
\centerline{\includegraphics[width=0.7\textwidth]{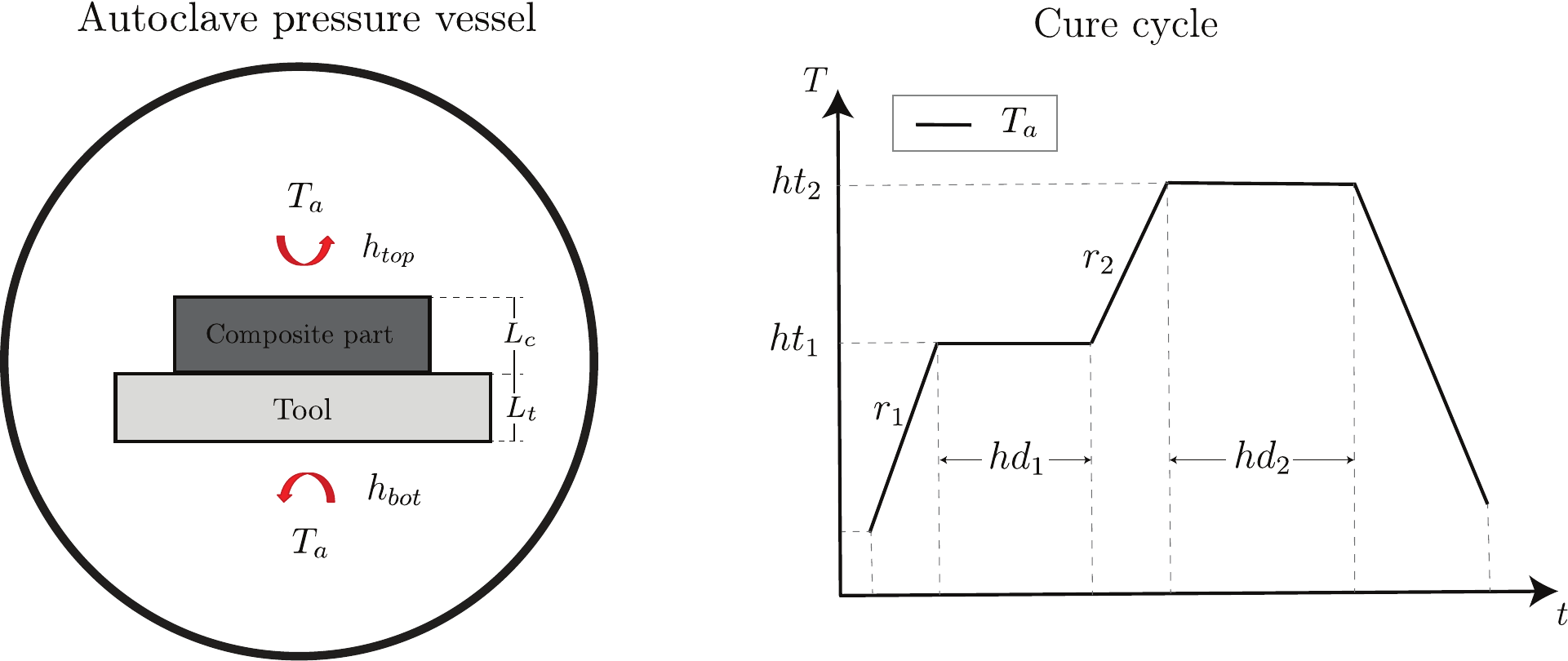}}
\caption{Schematic of Autoclave pressure vessel and typical two-hold cure cycle recipe for autoclave air temperature $T_a$ in composites manufacturing.}
\label{Curecycle_schematic}
\end{center}
\vskip -0.2in
\end{figure}

\subsection{Spatial local coordinates} \label{LocalCoordinates}

Figure \ref{localcoord} illustrates the local coordinates \(x_1\) and \(x_2\),  which are defined to manage inconsistencies in the total length of the system and the interface location resulting from the varying part and tool thicknesses. Substituting the local coordinates into the heat conduction, heat convection, and continuity equations results in the following transformed governing equations:

\begin{equation}
\left\{ \begin{aligned} 
  \frac{\partial T_t}{\partial t} &= \frac{a_t}{L_t^2}\frac{\partial^2T_t}{\partial x_1^2} \ \ \ \ \ \ \ \ \ \ \ \ \ \ \ \  \ \ \  x_1\in [0, 1]
  \\
  \frac{\partial T_c}{\partial t} &= \frac{a_c}{L_c^2}\frac{\partial^2T_c}{\partial x_2^2}+ b_c\frac{\partial \alpha}{\partial t} \ \ \ \ \  x_2\in [0, 1]
\end{aligned} \right.
\end{equation}

\begin{equation}
\begin{aligned}
\frac{\partial T_c}{\partial x_2}\mid_{x_2=1} &= \frac{h_{top} L_c}{k_c}(T_a-T_c\mid _{x_2=1})
\\
\frac{\partial T_t}{\partial x_1}\mid_{x_1=0} &= \frac{h_{bot} L_t}{k_t}(T_t\mid_{x_1=0} - T_a)
\\
T_t\mid_{x_1=1} &= T_c\mid_{x_2=0}
\\
\frac{k_t}{L_t}\frac{\partial T_t}{\partial {x_1}}\mid _{x_1=1} &= \frac{k_c}{L_c}\frac{\partial T_c}{\partial {x_2}}\mid _{x_2=0}.
\end{aligned}
\end{equation}

The use of local coordinates ensures identical coordinate domain size across all bi-material systems selected for training and testing. The thickness variation is then appropriately accounted for within the parameters of the differential equations (i.e., via the presence of \(L_c\) and \(L_t\) in such equations). This also enables treating the composite part and tool as standalone systems trained on separate networks with input variables normalized to 0 and 1. Specifically, as discussed in Section \ref{decoupledsec}, two DeepONets are allocated to capture the thermal behaviors of the part and tooling separately. While the individual DeepONets are responsible for learning separate systems, they also need to simultaneously satisfy the continuity conditions at the interface of the two materials. This is accomplished by defining two additional loss terms:
\begin{equation}
\begin{aligned}
    \mathcal{L}_{CT_1}(\theta_t, \theta_c) &= \frac{1}{NQ_{ct}}\sum_{i=1}^{N}\sum_{j=1}^{Q_{ct}}|G^{T_t}_{\theta_t}(u^{(i)})(y^{(i)}_j))-G^{T_c}_{\theta_c}(u^{(i)})(y^{(i)}_j))|^2
    \\
    \mathcal{L}_{CT_2}(\theta_t, \theta_c) &= \frac{1}{NQ_{ct}}\sum_{i=1}^{N}\sum_{j=1}^{Q_{ct}}|\frac{k_t}{L_t}\frac{\partial G^{T_t}_{\theta_t}(u^{(i)})(y^{(i)}_j))}{\partial x_1}-\frac{k_c}{L_c}\frac{\partial G^{T_c}_{\theta_c}(u^{(i)})(y^{(i)}_j))}{\partial x_2}|^2.
\end{aligned}
\end{equation}

Here \(\theta_t\) and \(\theta_c\) denote the weights associated with the part DeepONet \(G^{T_t}\) and tool DeepONet \(G^{T_c}\), respectively (see Figure \ref{EnPIDON}). \(Q_{ct}\) is the number of residual points evaluated at the materials’ interface.

\subsection{Training procedure} \label{Trainingprocess}
During the training phase, \(G^{T_t}\) is responsible for enforcing the tool’s IC,  bottom surface BC, and tool’s PDE. Similarly, \(G^{T_c}\) ensures the part's IC, top surface BC, and the part's PDE are satisfied. Additionally, the continuity conditions between the tool and part are maintained by jointly updating the weights of \(G^{T_t}\) and \(G^{T_c}\). Similarly, ODE loss (resin cure kinetics) is minimized by the sequential co-training of \(G^{T_c}\) and \(G^{\alpha}\). Furthermore, \(G^{\alpha}\) is tasked with minimizing the initial condition loss associated with DoC. We implemented the sequential learning approach \cite{niaki2021physics} to train the operators. The training began with updating the weights of \(G^{T_c}\) and \(G^{T_t}\) through their associated loss terms for 10 epochs while keeping the \(G^\alpha\)’s weights constant. Subsequently, \(G^\alpha\) was trained for 10 epochs while the other two operators remained frozen. The training ends after repeating this procedure 10 times.

\begin{figure}[h]
\vskip 0.2in
\begin{center}
\centerline{\includegraphics[width=0.8\textwidth]{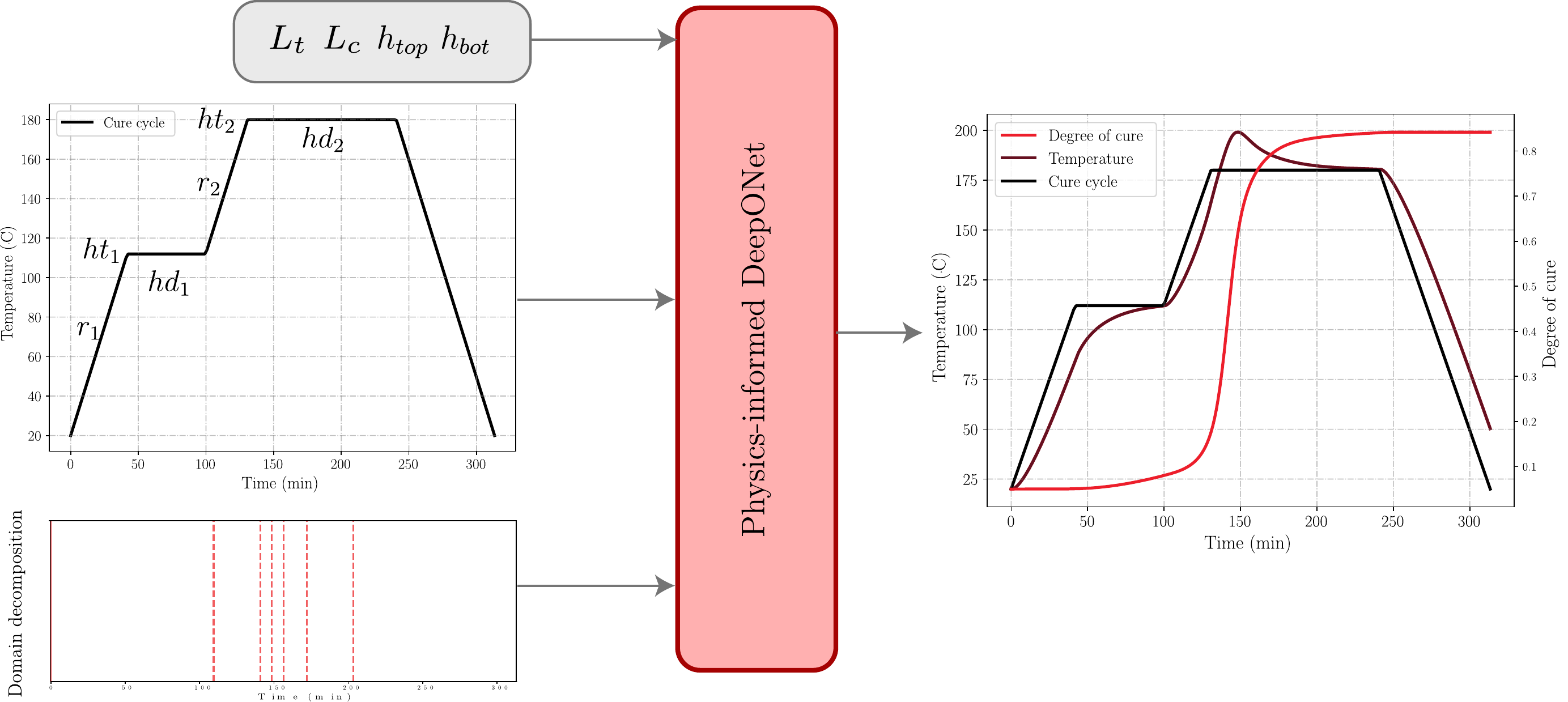}}
\caption{Proposed PIDON input-output functions mapping. PIDON takes in time-dependent (cure cycle) and time-independent ($L_t, L_c, h_{top}, h_{bot}$) functions as input and outputs temperature and DoC fields. The arrangement of temporal subdomains is also presented.}
\label{pidon_mapping}
\end{center}
\vskip -0.2in
\end{figure}

The input of BN2 is the sensory information of the air profile (i.e., cure cycle) surrounding the composite system during the curing process which enforces the boundary conditions. Various cure cycles with different numbers of isothermal holds as well as different heat ramp rates and hold durations are considered. For each cure cycle, the air temperature is recorded at 100 specified time steps (i.e., sensor locations) and the data is passed to BN2. BN1 on the other hand is fed with the remaining time-invariant process parameters. In this study, four process parameters including the top HTC, bottom HTC, tool’s thickness, and composite part’s thickness are considered (Figure \ref{pidon_mapping}). Table \ref{designRange} summarizes the design parameters and their corresponding ranges used for training the PIDON models in this study.

\begin{table}[h]
\caption{Design variables (input functions) and their corresponding ranges for three different design space sizes.}
\label{designRange}
\vskip 0.15in
\begin{center}
\begin{small}
\begin{sc}
\begin{tabular}{lllll}
\toprule
Design    & Description   & \multicolumn{3}{c}{Design space size} \\
parameter &  &  Small       & Medium       & Large    \\
\midrule
$h_{top}$  \((\textup{W/m\textsuperscript{2} K})\)  &   Top HTC    & [90, 120]  & [80, 120]   & [70, 120] \\
$h_{bot}$  \((\textup{W/m\textsuperscript{2} K})\) &   Bottom HTC      & [60, 90]  & [50, 90]   & [50, 100]\\
$r_1$  \(\textup{(°C/min)}\)  &   Ramp 1    & [1.9, 2.8]  & [1.7, 3]   & [1.5, 3] \\
$ht_1$ \(\textup{(°C)}\)    &   Hold 1 temperature      & [110, 115]  & [105, 115]   & [105, 120]\\
$hd_1$ \(\textup{(min)}\)    &   Hold 1 duration    & [55, 63]  & [52, 63]  & [50, 65]\\
$r_2$  \(\textup{(°C/min)}\)   &   Ramp 2      & [1.9, 2.8]  & [1.7, 3]   & [1.5, 3]\\
$ht_2$  \(\textup{(°C)}\)   &   Hold 2 temperature    & [178, 183]  & [175, 185]   & [170, 185] \\
$hd_2$ \(\textup{(min)}\)   &   Hold 2 duration      & [105, 115]  & [105, 120]   & [105, 120]\\
$L_t$ \(\textup{(cm)}\)   &   Tool thickness    & [2, 3.5]  & [2, 4]   & [2, 5] \\
$L_c$ \(\textup{(cm)}\)  &   Part thickness      & [2.5, 3.5]  & [2.5, 3.5]   & [2.5, 3.5]\\

\bottomrule
\end{tabular}
\end{sc}
\end{small}
\end{center}
\vskip -0.1in
\end{table}

\section{DeepONet} \label{deeponetAppx}

\begin{figure}[ht]
\vskip 0.2in
\begin{center}
\centerline{\includegraphics[width=0.7\textwidth]{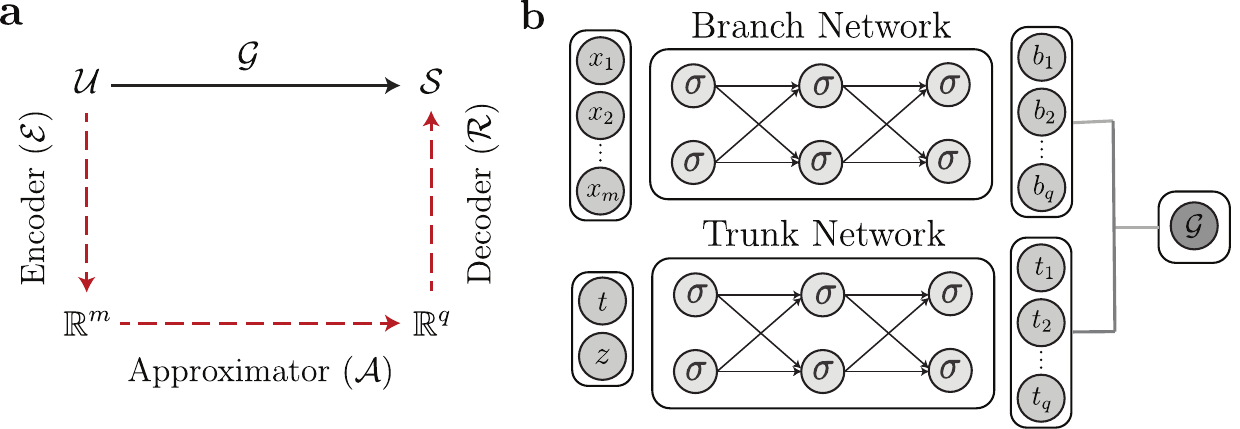}}
\caption{Schematic of DeepONet decomposition into Encoder, Approximator, and Decoder (a); architecture of vanilla DeepONet with a branch net, a trunk net, and a linear decoder.}
\label{fig1}
\end{center}
\vskip -0.2in
\end{figure}

Based on the universal approximation theorem for operators, \cite{chen1995universal} proposed \textit{operator nets}, a neural network architecture that approximates nonlinear operators that map infinite dimensional Banach spaces. An operator net consists of two shallow neural networks, namely, branch net and trunk net, which encode the input functions and system coordinates, respectively. The branch and trunk nets are merged to approximate the underlying operator solution. \cite{lu2021learning} proposed a more expressive variant of the operator net named DeepONet by replacing shallow networks with deep neural networks. The architecture of the DeepONet can naturally be decomposed into three main components: an encoder, an approximator, and a decoder, as illustrated in  Figure \ref{fig1}.a \cite{lanthaler2022error}. The encoder is responsible for mapping the infinite-dimensional input space to a finite-dimensional space. This is crucial for training the operator as the input functions must be expressed discretely to implement the network approximations. In other words, the continuous input functions are mapped to their discretized representation (i.e., finite-dimensional space) by pointwise evaluations at \(m\) fixed sensor points \(x_j\). The approximator is parameterized by a deep neural network that maps the sensor point evaluations \(u = [u(x_1), u(x_2), ..., u(x_m)]\) to a finite-dimensional feature representation \(b = [b_1, b_2, ..., b_q]^T \in R^q\). The composition of the encoder and approximator results in the branch net of DeepONet expressed by \(\beta(u)= \mathcal{A} \circ \mathcal{E}(u) \). Similar to the branch net, the trunk net is parameterized by a deep neural network that encodes the inputs of the PDE system \(y\) to a feature embedding \(t = [t_1, t_2, ..., t_q]^T \in R^q\) with the same size as the branch net’s output (Figure \ref{fig1}.b). Finally, the decoder takes the output of the branch (\(q\) coefficients) and trunk nets (\(q\) basis functions) and calculates the DeepONet’s output using an element-wise product operation followed by a summation, \(G_\theta(u)(y) = \sum_{k=1}^{q} b_kt_k + b_0\). The bias term \(b_0\) is added in practice to improve the generalization performance of DeepONet \(G\). The decoder can also be seen as a single network (trunk net) with its weights in the last layer parameterized by another network (branch net). In a supervised learning fashion, the DeepONet can be trained by minimizing the error between the model’s predicted output and the actual operator solution across a range of training input functions.

\section{Additional results}

\begin{figure}[ht]
\vskip 0.2in
\begin{center}
\centerline{\includegraphics[width=0.9\textwidth]{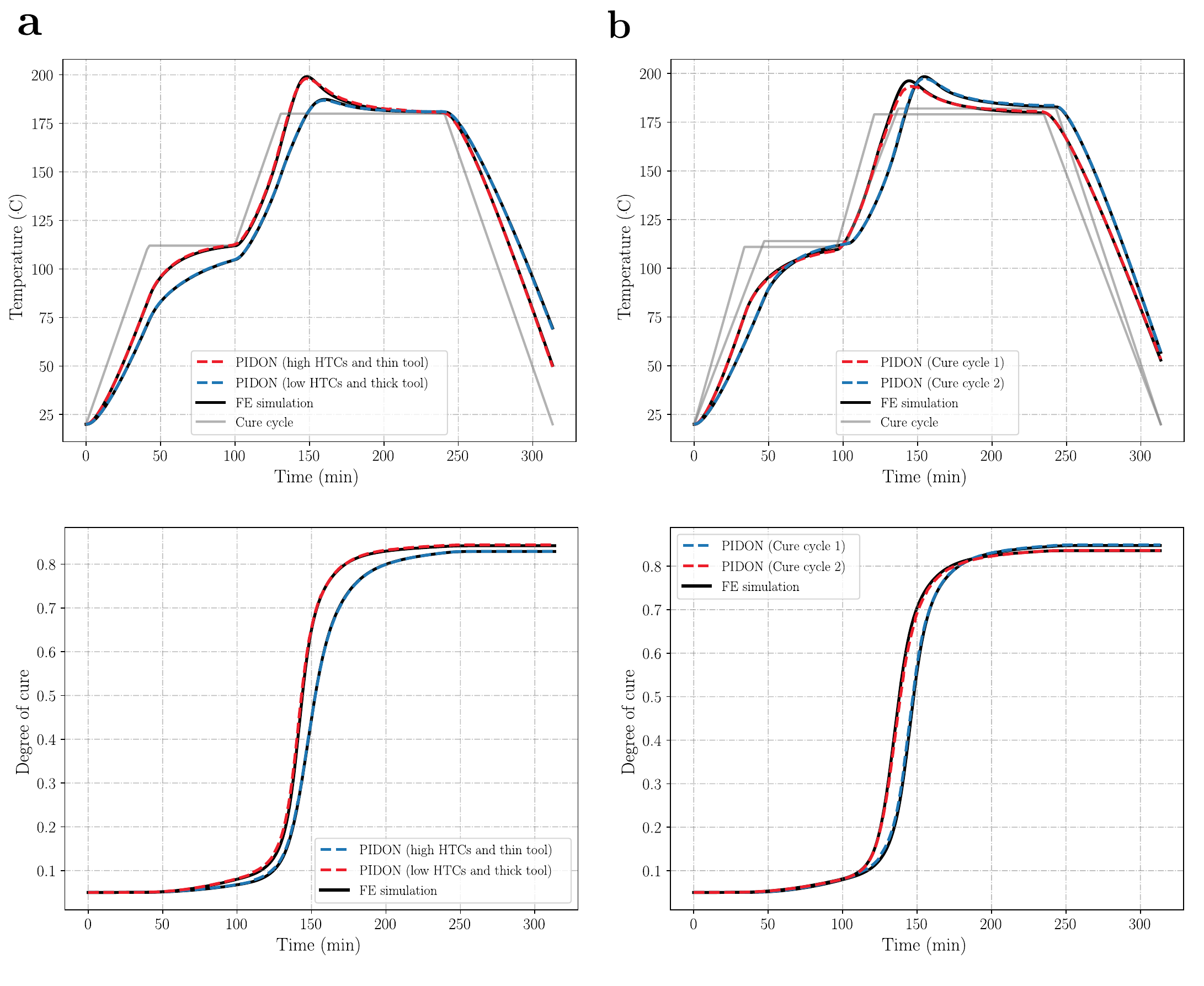}}
\caption{Zero-shot prediction performance of PIDON at composite part mid-point for: a) high HTCs and thin tool vs. low HTCs and thick tool; b) two different cure cycles (shown in gray.)}
\label{morepred}
\end{center}
\vskip -0.2in
\end{figure}

\begin{figure}[ht]
\vskip 0.2in
\begin{center}
\centerline{\includegraphics[width=\textwidth]{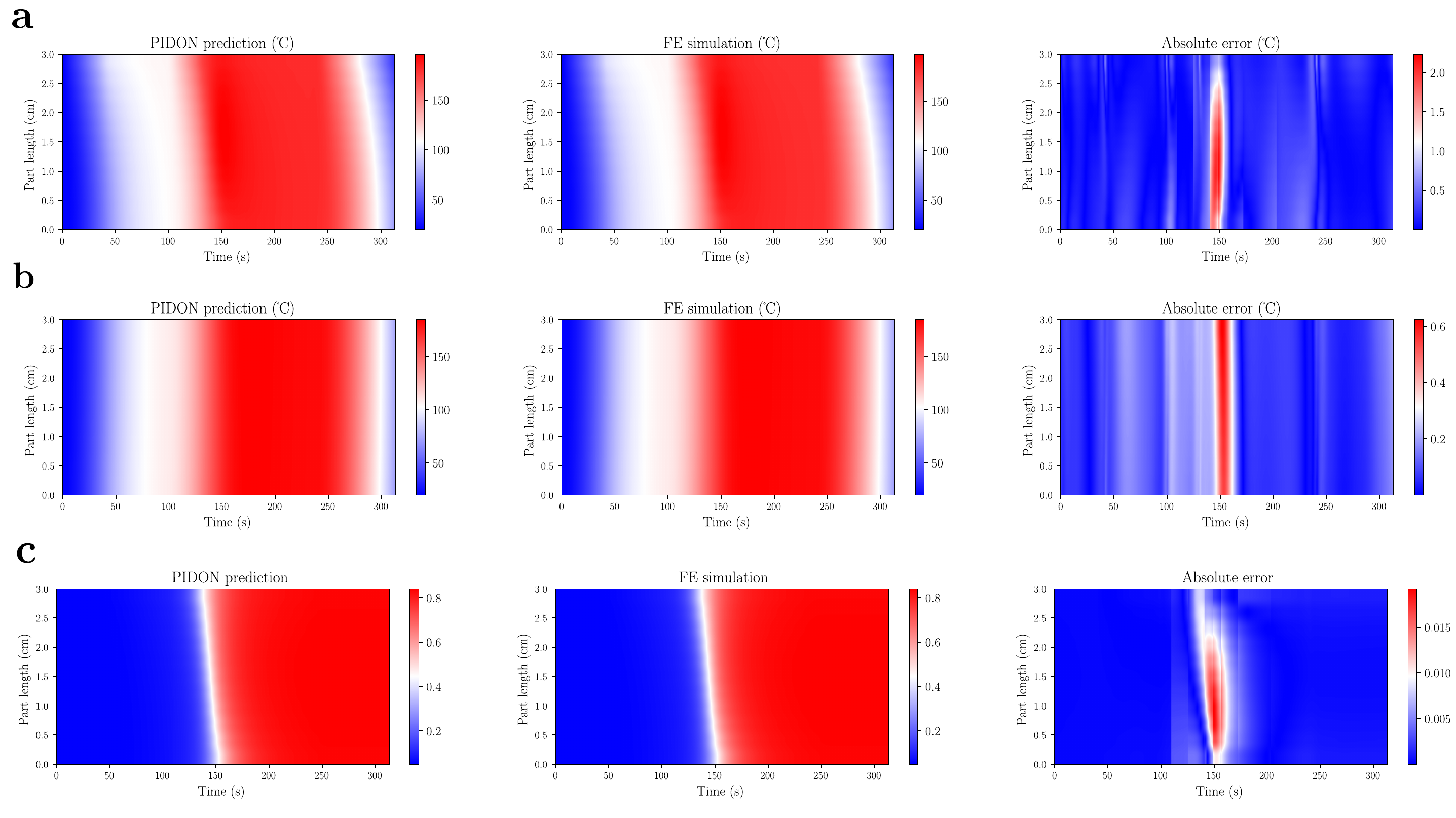}}
\caption{Comparison of PIDON prediction and FE simulation for Part temperature (a), tool temperature (b) and DoC (c) for a test case with the design variables: \(h_{top} = 75, h_{bot} = 115, r_1=r_2=2.2, ht_1= 110, hd_1 = 58, ht_2 = 180, hd_2 = 105, L_t = 0.025, L_c = 0.03\). }
\label{imshow}
\end{center}
\vskip -0.2in
\end{figure}

\begin{figure}[ht]
\vskip 0.2in
\begin{center}
\centerline{\includegraphics[width=\textwidth]{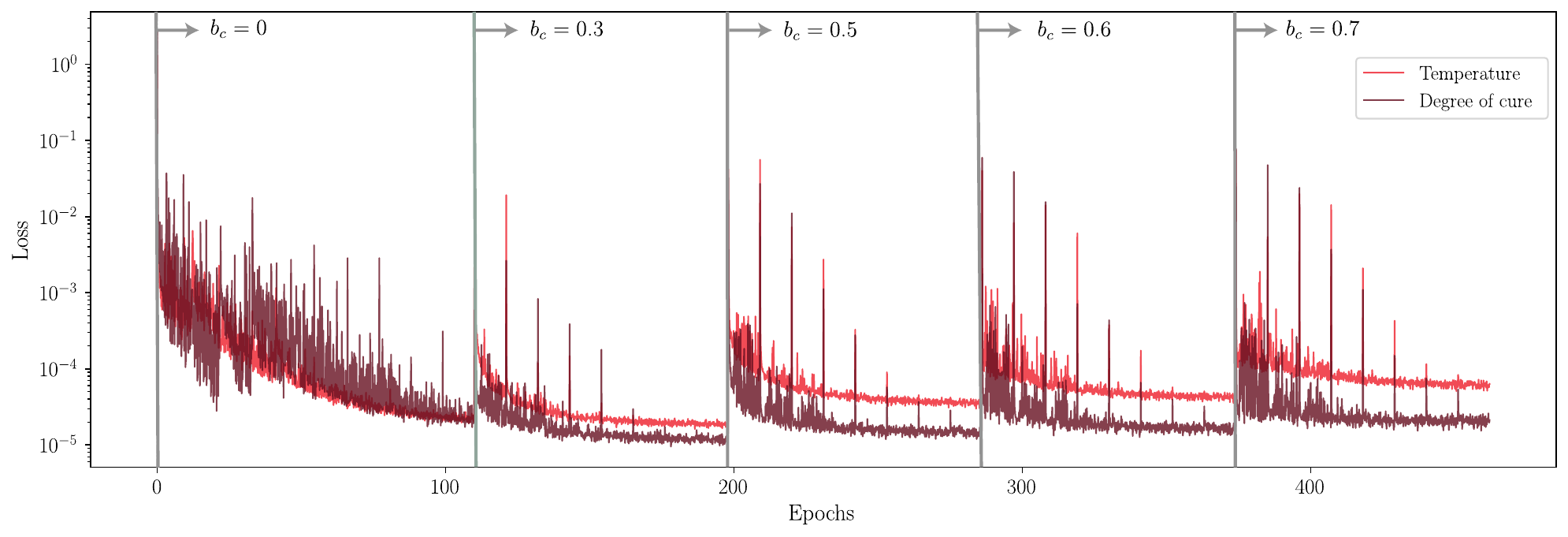}}
\caption{The evolution of temperature and DoC training losses during curriculum learning at various values of heat generation coefficient \(b_c\).}
\label{CLprogress}
\end{center}
\vskip -0.2in
\end{figure}

\end{document}